\documentclass{article}
\usepackage{ijcai15}
\usepackage{times}

\usepackage{epsfig}
\usepackage{graphicx}
\usepackage{amsmath}
\usepackage{amssymb}
\usepackage{dsfont}
\usepackage{subfigure}
\usepackage{multirow}
\usepackage{url}
\usepackage{balance}
\usepackage{sidecap}
\usepackage{array}

\newcolumntype{V}{>{$\vcenter\bgroup\hbox\bgroup}c<{\egroup\egroup$}}
\def\Hline{\noalign{\hrule height 4\arrayrulewidth}}

\graphicspath{{./imgs/},{../2015_PR1/img-j/}}

\title{Groupwise Registration of Aerial Images}
\author{
    Ognjen Arandjelovi\'c$^\dag$ , Duc-Son Pham$^\ddag$ , and Svetha Venkatesh$^\dag$ \\
    \begin{tabular}{cc}
      $^\dag$ Centre for Pattern Recognition \& Data Analytics & $^\ddag$ Department of Computing\\
      ~~~Deakin University, Australia                             & ~~~Curtin University, Australia
    \end{tabular}
}

\begin{document}

\maketitle

\begin{abstract}
  This paper addresses the task of time separated aerial image registration. The ability to solve this problem accurately and reliably is important for a variety of subsequent image understanding applications. The principal challenge lies in the extent and nature of transient appearance variation that a land area can undergo, such as that caused by the change in illumination conditions, seasonal variations, or the occlusion by non-persistent objects (people, cars). Our work introduces several novelties: (i) unlike all previous work on aerial image registration, we approach the problem using a set-based paradigm; (ii) we show how local, pair-wise constraints can be used to enforce a globally good registration using a constraints graph structure; (iii) we show how a simple holistic representation derived from raw aerial images can be used as a basic building block of the constraints graph in a manner which achieves both high registration accuracy and speed. We demonstrate: (i) that the proposed method outperforms the state-of-the-art for pair-wise registration already, achieving greater accuracy and reliability, while at the same time reducing the computational cost of the task; and (ii) that the increase in the number of available images in a set consistently reduces the average registration error.
\end{abstract}

\section{Introduction}\label{s:intro}
\begin{figure}[thp]
  \centering
  \subfigure[]{\includegraphics[width=0.13\textwidth]{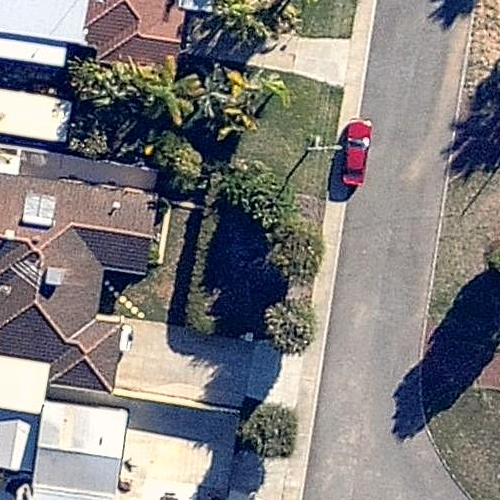}}~~
  \subfigure[]{\includegraphics[width=0.13\textwidth]{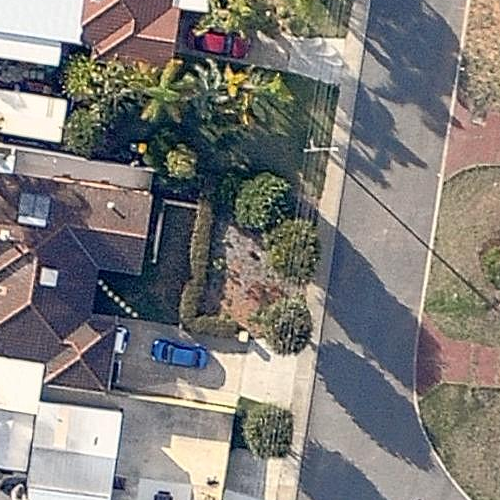}}~~
  \subfigure[]{\includegraphics[width=0.13\textwidth]{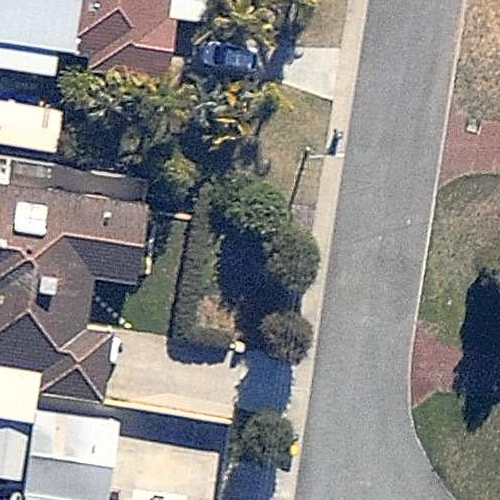}}
  \subfigure[]{\includegraphics[width=0.13\textwidth]{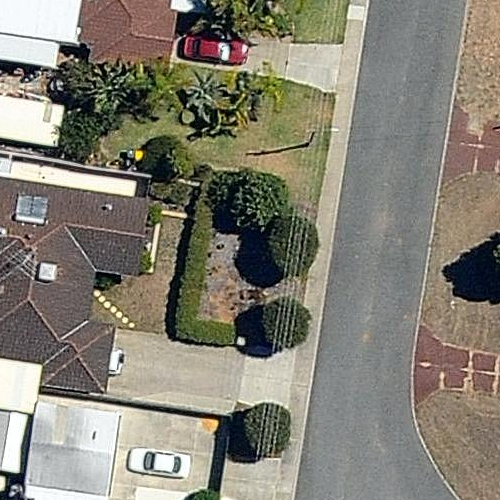}}~~
  \subfigure[]{\includegraphics[width=0.13\textwidth]{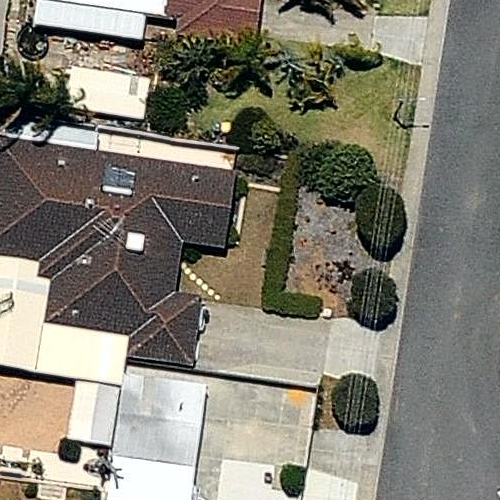}}~~
  \subfigure[]{\includegraphics[width=0.13\textwidth]{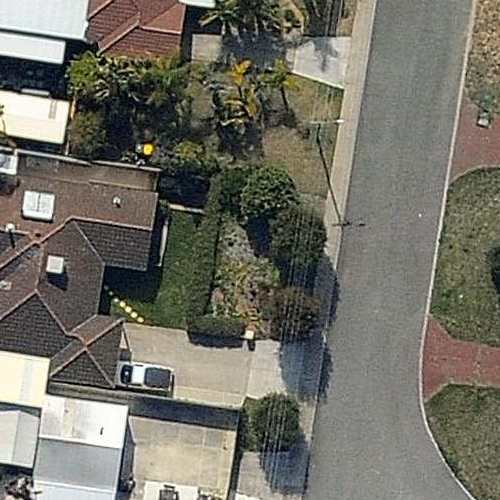}}
  \caption{\small Input images (for easier visualization, the patches shown include only approximately 10\% of the area used as actual input). These correspond to approximately the same land area imaged on different days and different times of the day, and registered using a state-of-the-art commercial registration system which uses both GPS and image data. Substantial registration errors remain (e.g.\ the misalignment between (d) and (e) is 78 pixels).}
  \label{f:patches}
\end{figure}

The goal of the present work is to achieve accurate registration of time separated aerial images. The key challenge of this task emerges as a consequence of the potentially large transient appearance changes, such as those which may be caused by different illumination conditions, seasonal variations, or mobile objects with a non-permanent presence (e.g.\ people, cars, lawnmowers). Some of these challenges are illustrated in Fig~\ref{f:patches}, using a sample of images taken from our evaluation data set. Reliable registration is an important pre-processing step required by a  wide range of practical applications including semantic labelling of images \cite{MnihHint2012}, and the detection of meaningful (high-level), structural changes \cite{Chha2009,LuoLi2011}. Unlike previous work on aerial images, we formulate and address the registration problem using an image set-based framework, rather than a sequence of independent pair-wise registrations.

Registration, as a general problem of geometric normalization, is pervasive in computer vision. Unsurprisingly, the corpus of relevant previous work is rich and varied, often with a high degree of domain specificity~\cite{ZitoFlus2003}. In aerial imaging applications, most registration approaches described in the literature typically focus on man-made structures, \textit{a priori} choosing to exploit the presence of line features \cite{WongClau2007}, rectangular buildings \cite{NoroNeva2001}, or roads \cite{MnihHint2010}. All of these methods register images in a pair-wise manner, either aligning an aerial image to an aerial image, or an aerial image to a map. No previous work on aerial image registration operates directly on image sets as input, nor is readily extended to this problem setup.

To place the present work in broader context and better appreciate our contributions, it is worth noting that set-based registration methods have been described in other application domains of computer vision, most notably in the field of medical image understanding \cite{MetzKleiSchavanW+2011}. However, an examination of these approaches shows that they too cannot be readily applied on the problem we consider in this paper. Indeed, medical images are often taken in calibrated conditions, consistent across acquisitions, while aerial images exhibit extreme variations due to uncontrollable illumination and seasonal effects, amongst others. For example, the set-based method based on Havrda-Carv{\'a}t cumulative residual entropy proposed in \cite{ChenVemuRangEise2010} requires the shapes of objects of interest to be known in advance. The approaches described in \cite{LordHoVemu2007} and \cite{LiManjMitr1995} suffer from a similar limitation in this context, given that they require reliable contour information. Their target domain being void of such challenges, in \cite{WachNava2012} the authors do not consider the issues of illumination change or the potential presence of transient structures and occlusions. Similar observations hold for a number of other approaches \cite{ThevRuttUnse1998,ForoZeruBert2002}. A major recent development draws from the advances in sparsity learning, often with the specific focus on the alignment of images of faces \cite{WagnWrigGaneZhou+2012,GhosManj2013}. Because of their computational cost, these methods are limited in their application to images prohibitively small for our problem. In actual practice, the standard registration procedure used by geographers involves the manual selection of a set of placemarks, followed by the application of a bundle adjustment algorithm \cite{TrigMcLaHartFitz2000}. This is not only a laborious process but also one which requires considerable expertize and experience because the number and the choice of placemark locations is highly scene specific yet crucial for the success of the overall scheme.

Our work introduces several major novelties of significance. Firstly, we show how a simple quasi-invariant representation derived from raw aerial images, employed in a coarse-to-fine fashion with a suitable matching function can be used to improve the accuracy and reliability of registration at the level of pair-wise image alignment already. Secondly, we describe how this approach can be employed as a part of a novel set-based registration framework. The proposed framework is built upon what we term a constraints graph, which is used to propagate local, pair-wise registration information across the image set. We demonstrate that the resulting method outperforms the current state-of-the-art in aerial image registration both in terms of accuracy and reliability, as well as speed.

\section{Proposed approach}
In this section we describe the key technical contributions of the present paper. We start with an overview of the proposed approach and its key constituent elements, and follow with a detailed description of each.

\subsection{Overview}
At the centre of the proposed approach is an optimization scheme. While the optimization scheme itself is global (i.e.\ it operates jointly over the entire image set), globality is achieved through the propagation of local registration constraints. The aim is to co-register an entire image set by finding the best compromise between different pair-wise registrations. Both for the sake of computational efficiency as well as reasons inherently linked to the nature of the problem under consideration (see Sec~\ref{ss:constraints} for detail), only some pair-wise registrations contribute to the objective (fitness) function which is maximized. Like most previous work, we constrain our consideration to translation only.

\subsection{Constraints graph}\label{ss:constraints}
The method we propose in this paper is founded on two key ideas. The first of these is that the registration of an entire set of images can be assessed and should be formulated as a function of pair-wise registrations. The second idea and premise, is that the magnitude and nature of confounding variation present in realistic images of approximately the same geographical location acquired make pair-wise registration difficult and unreliable. Thus, our method aims to find the best solution (relative registration adjustment) which balances different pair-wise assessments of registration quality. Formally, our registration can be written as an optimization problem which comprises the maximization of the following fitness function (we use the term ``fitness function'' to emphasize that the desired solution maximizes its value, rather than ``objective function'' which is less specific and is generally used when minimization is sought):
{\small\begin{align}
  J(\{ \Delta r_k \};& \{ I_k \})=
    \sum_{i=1}^n \sum_{j=1}^n \Big\{ w_{i,j} \times \hat{\rho}(\Delta r_i-\Delta r_j; \zeta(I_i), \zeta(I_j)) \Big\}.
  \label{e:fitfn}
\end{align}}
Here, $J(\{ \Delta r_k \}; \{ I_k \})$ is the value of the fitness function for the set of registration adjustments $\{ \Delta r_k \}$ relative to a specific reference image (the choice of the reference image does not affect the result so we consistently select the first image of the set, i.e.\ $I_1$, as the reference image) for the input image set $\{I_k\}=\{I_1,\ldots,I_n\}$, $\zeta(I_i)$ the quasi-invariant representation of the view captured by image $I_i$, $\hat{\rho}(\Delta r_i - \Delta r_j;\zeta_1,\zeta_2)$ a measure of pair-wise registration agreement between quasi-invariant representations $\zeta_1$ and $\zeta_2$ geometrically adjusted by $\Delta r_i-\Delta r_j$, and $w_{i,j}$ a binary weight (valued 0 or 1) which includes or excludes the contribution of the corresponding ``local'' i.e.\ pair-wise registration to the global criterion.

The design of the local elements in \eqref{e:fitfn} -- specifically the quasi-invariant image representation and the corresponding distance measure -- is addressed in Sec~\ref{ss:local}. Presently we focus on the global issues, that is, the problem of selecting which weights $w_{i,j}$ should vanish and which should assume a unitary value. We think of this process as the construction of a constraints graph $G=(V,E)$ whose vertices (nodes) correspond to input images $V =\{I_1,I_2,\ldots,I_n\}$ and whose edges $E =\{w_{1,1},w_{1,2},\ldots,w_{n,n-1},w_{n,n}\}$ encode the set of local constraints which contribute to the set-based registration fitness function.

Considering that we assume that the extent and the nature of appearance changes across input images present a major challenge to pair-wise registration, it is a premise inherent in our approach that the additional constraints and information should come from the structure of the graph $G$. Thus the central question becomes what the topology of this graph should be to ensure that additional information is indeed extracted.

Qualitatively speaking, we identify two types of good registration reinforcement that our constraints graph can effect. The first of these concerns similar input images and thus acts in a proximal fashion in the image space. The intuition is that similar images, i.e.\ images which are close in the input image space in the Euclidean distance sense, correspond to the scene imaged in similar conditions (e.g.\ few differences due to transient objects in the images, similar illumination conditions etc). By virtue of this observation, such images should be easier to register in a pair-wise manner. Consequently, there should be a connection between the corresponding nodes in the constraints group so as to ensure that the global registration is built upon such reliable pair-wise registrations. This allows reliable registrations to propagate their initially local constraints across the graph, achieving global effect. The second type of constraint we identify acts in a rather opposite manner from the previously described proximal constraint, in that it seeks to connect images which are distant from others. Intuitively speaking, these are images which have been acquired in conditions very much unlike any of the other images (more generally, we can talk about cliques of distant images which are separated from the rest of the data). The reason why good connectivity of the graph nodes corresponding to these images is desirable stems from the observation that these images cannot be included in the set registration scheme by means of concatenated reliable pair-wise registrations -- some other means of meaningful polling of information from the rest of the graph is needed. The premise behind the idea that distant images (or indeed cliques of such images) should be richly connected to the rest of the graph is that pair-wise appearance differences corresponding to their connection are likely to be approximately uncorrelated. By including a rich set of connections, the effect of changeable elements of the scene is outweighed by the persistent and reliable structures which remain stable across different connections.

\begin{figure}[t]
  \centering
  \subfigure[1: $k$-nearest n/bours]{\includegraphics[width=0.22\textwidth]{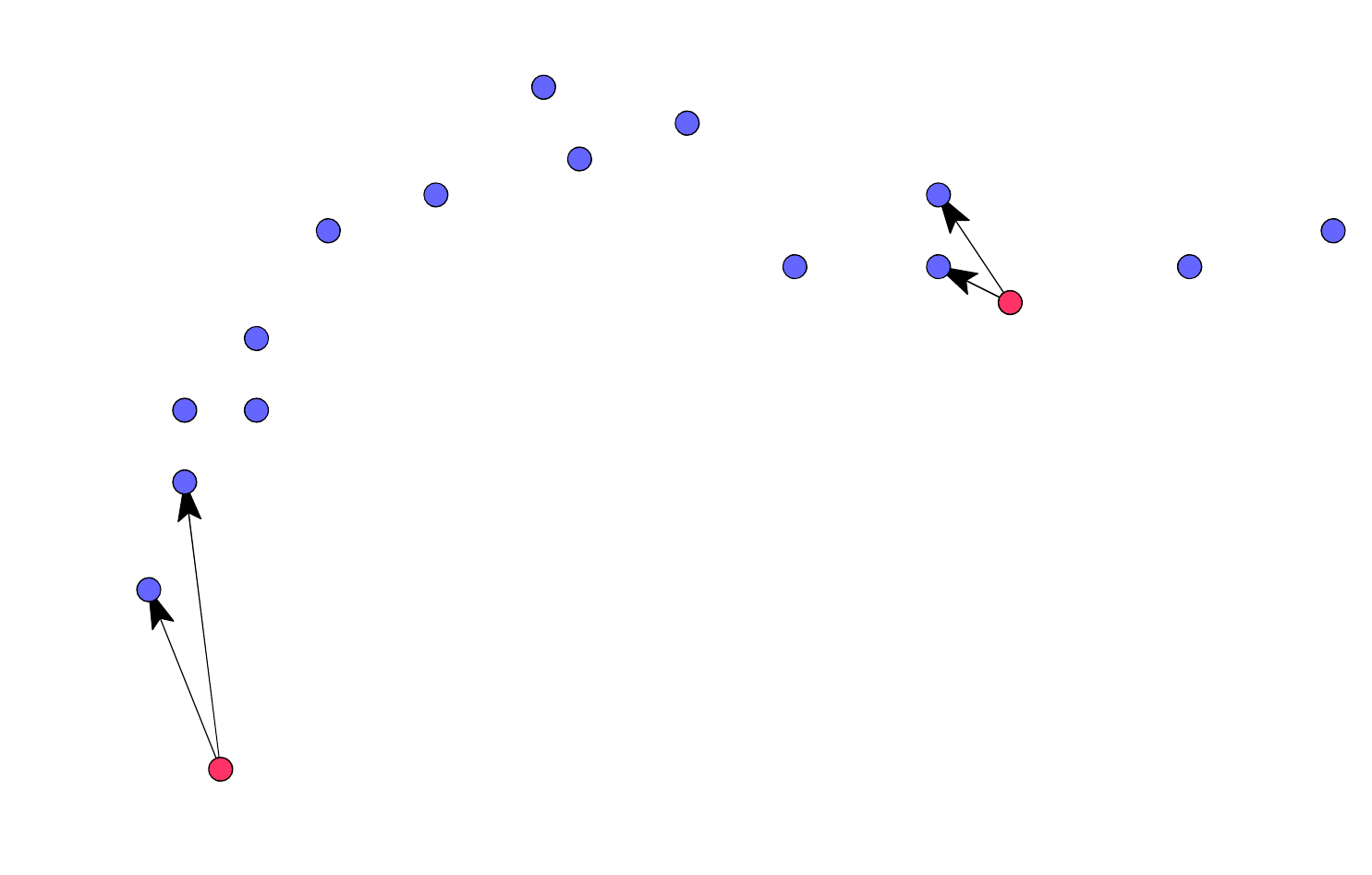}}\hspace{5pt}
  \subfigure[2: thresh.\ proximity]{\includegraphics[width=0.22\textwidth]{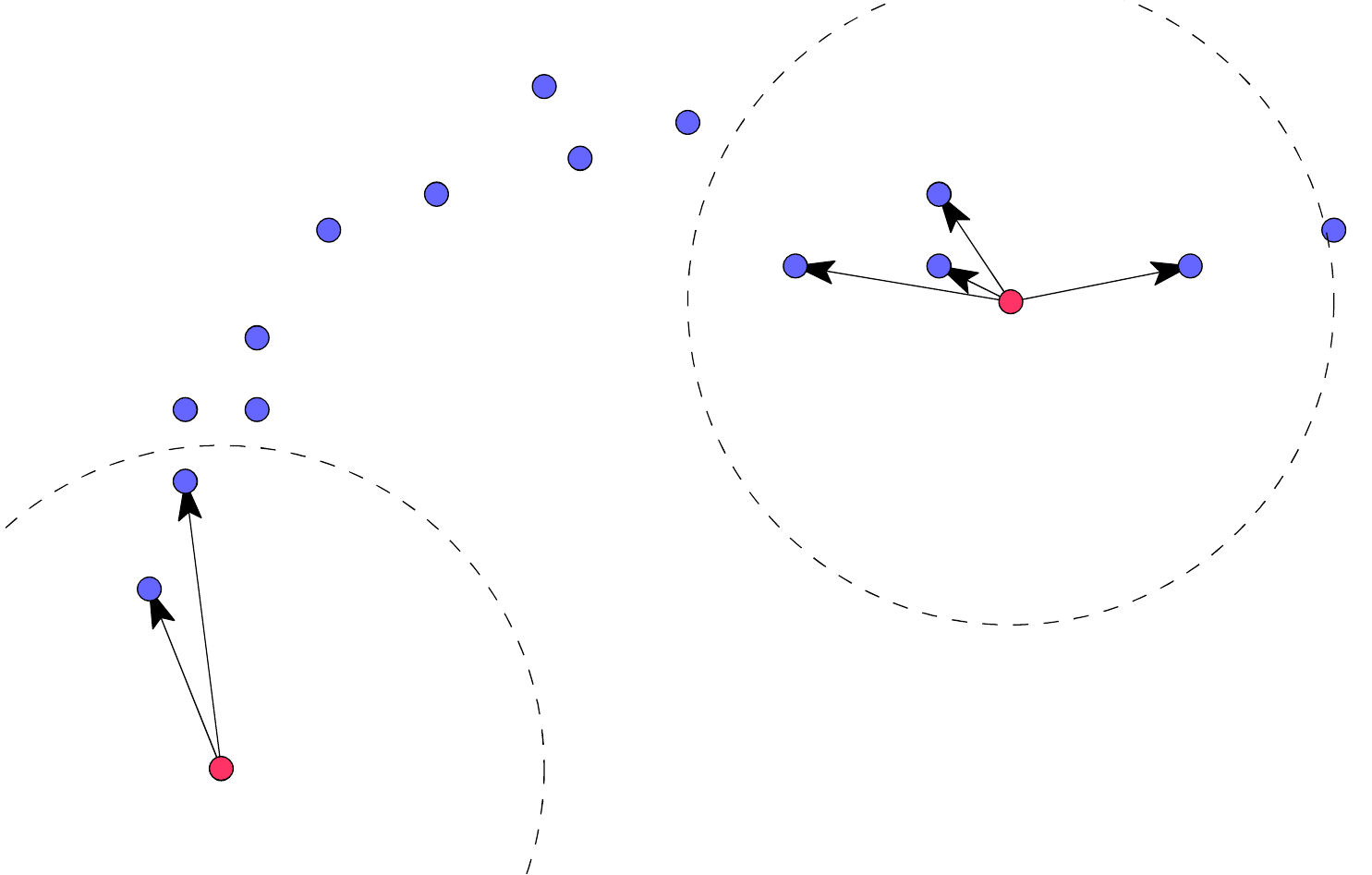}}
  \subfigure[3: $k$-furthest n/bours]{\includegraphics[width=0.22\textwidth]{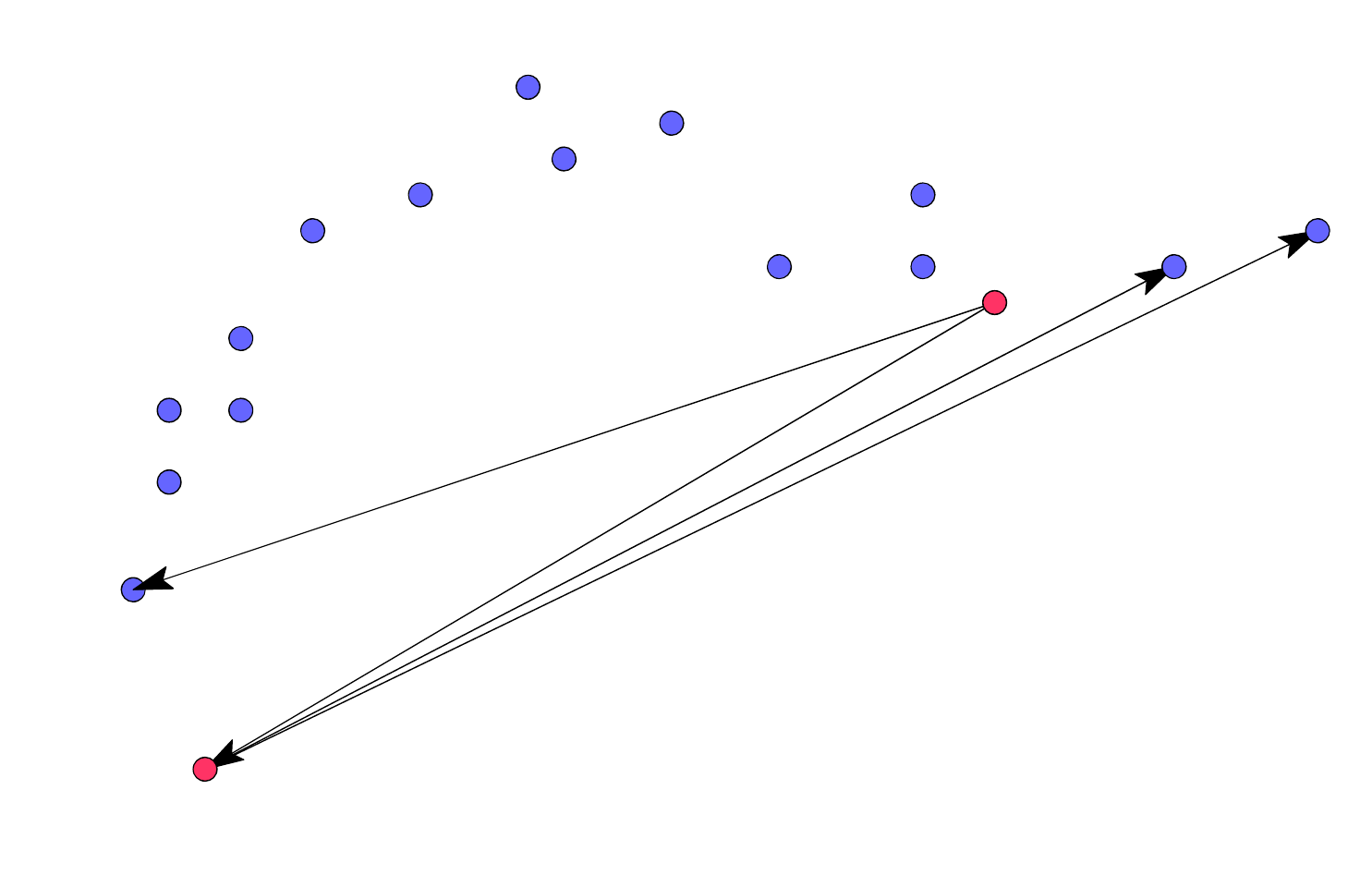}}\hspace{5pt}
  \subfigure[4: thresh.\ remoteness]{\includegraphics[width=0.22\textwidth]{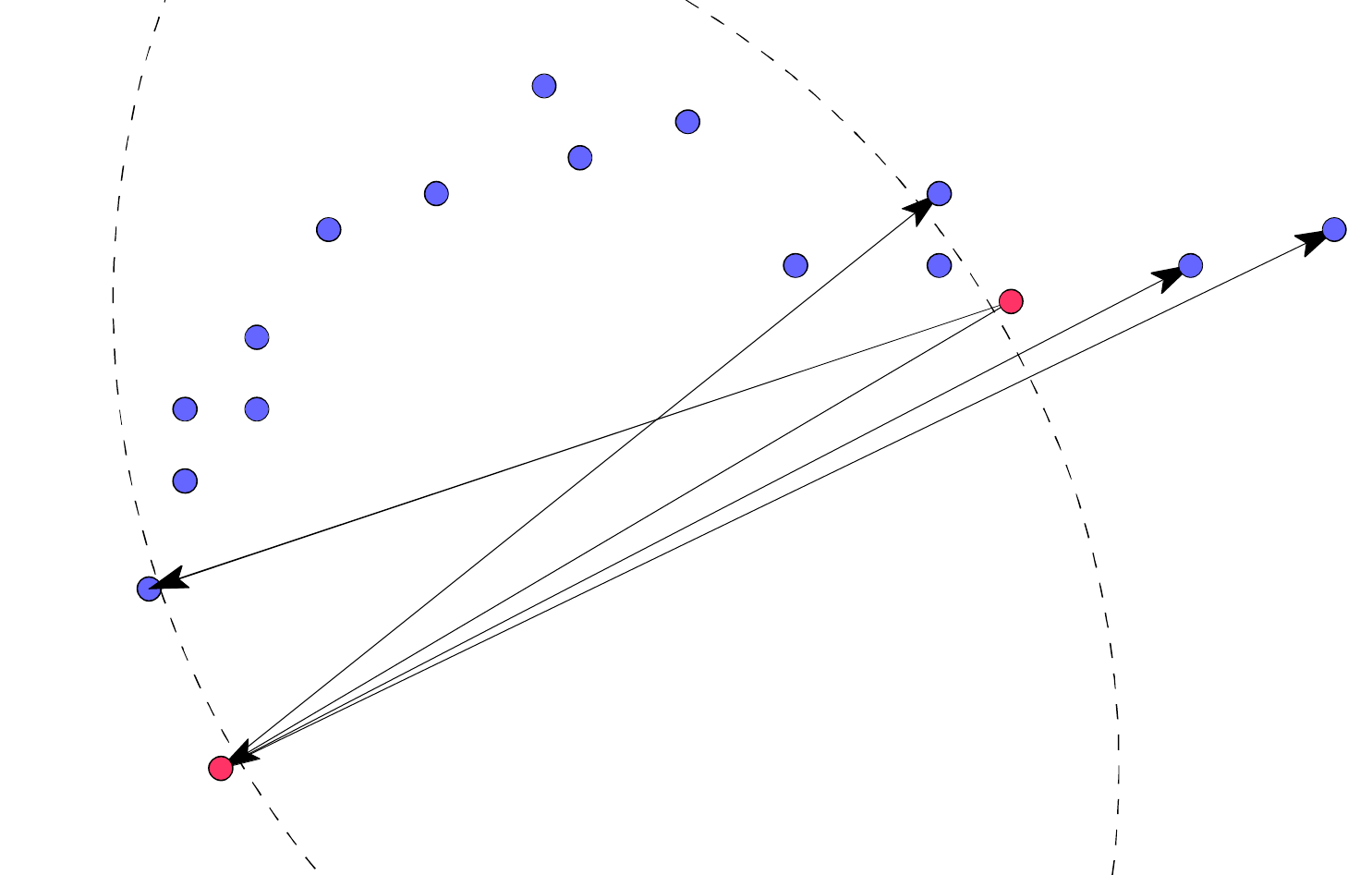}}

  \caption{\small Illustration of four schemes for constructing the local constraints graph over a set of images. Images (blue circles) are conceptually shown projected onto the 2D principal component space. }
  \label{f:schemes}
\end{figure}

Based on these two key ideas, we investigated four different elementary blocks -- building schemes -- used to construct the constraints graph. Two of them are used to establish proximal connections, while the other two are their distal analogues:
\begin{itemize}
  \item scheme 1: connections local in the Euclidean sense:
  {\small\begin{align}
    w_{i,j} =
    \begin{cases}
      0 ~:~ i=j ~\vee \text{ $I_j$ not one of $k$ nearest neighbours of $I_i$}\\
      1 ~:~ \text{$I_j$ is one of $k$ nearest neighbours of $I_i$}
    \end{cases}
  \end{align}}
  \item scheme 2: connections local in the Euclidean sense:
  {\small\begin{align}
    w_{i,j} =
    \begin{cases}
      0 ~:~ i=j ~\vee \text{ $d(I_i,I_j) > d_{thres1}$}\\
      1 ~:~ i\neq j ~\wedge \text{ $d(I_i,I_j) \leq d_{thres1}$}
    \end{cases}
  \end{align}}
  \item scheme 3: connections distal in the Euclidean sense:
  {\small\begin{align}
    w_{i,j} =
    \begin{cases}
      0 ~:~ i=j ~\vee \text{ $I_j$ not one of $k$ furthest images from $I_i$}\\
      1 ~:~ \text{$I_j$ is one of $k$ furthest neighbours of $I_i$}
    \end{cases}
  \end{align}}
  \item scheme 4: connections distal in the Euclidean sense:
  {\small\begin{align}
    w_{i,j} =
    \begin{cases}
      0 ~:~ \text{ $d(I_i,I_j) < d_{thres2}$}\\
      1 ~:~ \text{ $d(I_i,I_j) \geq d_{thres2}$}
    \end{cases}
  \end{align}}
\end{itemize}
where the distance between two images is in all cases measured in the original Euclidean space:
{\small
\begin{align}
  d_E(I_i,I_j) = \sqrt{\sum_r \Big[I_i(r)-I_j(r)\Big]^2.}
\end{align}}
Notice that schemes~2 and~4 result in symmetric edges (i.e.\ $w_{i,j}=1 \Leftrightarrow w_{j,i}=1$) while in general this is not the case for schemes~1 and~3. The four schemes are illustrated and compared conceptually in Fig~\ref{f:schemes}.

\begin{figure}[t]
  \centering
  \includegraphics[width=0.3\textwidth]{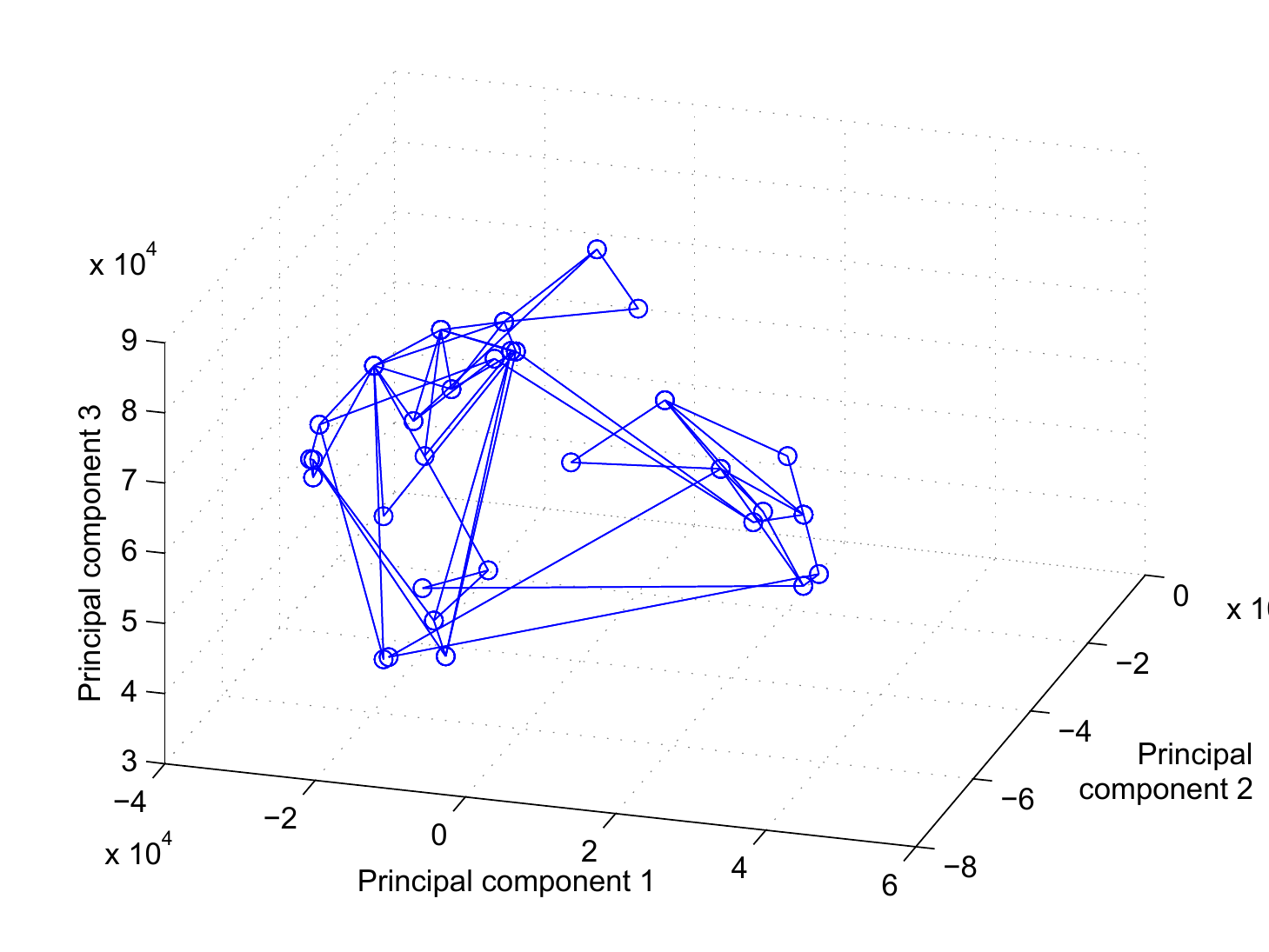}
  \caption{\small A constraints graph is used to propagate globally registration information from pair-wise image comparisons. Its nodes correspond to input images (here shown projected to their 3D linear principal subspace), while the connections between them encode which pair-wise comparisons contribute to the fitness function used to quantify the quality of registration on the level of the entire set. }
  \label{f:graph3PCA}
\end{figure}

We obtained the best results by combining two schemes, one proximal and one distal. The choice of the specific schemes was not found to affect the results significantly, and henceforth we adopt the combined use of schemes~1 and~3. An example of a graph built in this fashion is shown in Fig~\ref{f:graph3PCA}.

\subsection{Local constraint: pair-wise registration quality}\label{ss:local}
In the previous section we described how in our method `local', pair-wise registration information is propagated globally, that is, across the entire set of images being registered. The aim was to integrate the available information in a meaningful manner which leads to a globally good solution by virtue of local constraints. From this it is clear that ultimately the elementary building block of the scheme and the potential bottleneck is to be found in the way registration between a pair of images is assessed -- while the extent of the challenge posed by large appearance changes makes it unrealistic to expect a highly accurate result when only pairs of images are used, it is crucial that pair-wise registration is sufficiently powerful to drive the constraints graph optimization.

Our initial experiments with a variety of interest point detectors and local feature descriptors suggested that the extent of local appearance changes in our data is so substantial that very few reliable keypoint matches could be made~\cite{AranPhamVenk2015c}. Furthermore, we found that keypoints-based approaches did not readily lend themselves to an efficient integration in our constraints graph framework. Thus, we developed a holistic approach instead. Our approach on this, pair-wise level consists of two steps. Firstly, input images are processed using simple filters to produce a quasi-illumination invariant representation (the effects of illumination are readily recognized as effecting the most substantial changes between different images). The assessment of the registration quality for a particular translation is then quantified using the simple normalized cross-correlation coefficient.

\begin{SCfigure}
  \centering
  \includegraphics[width=0.28\textwidth]{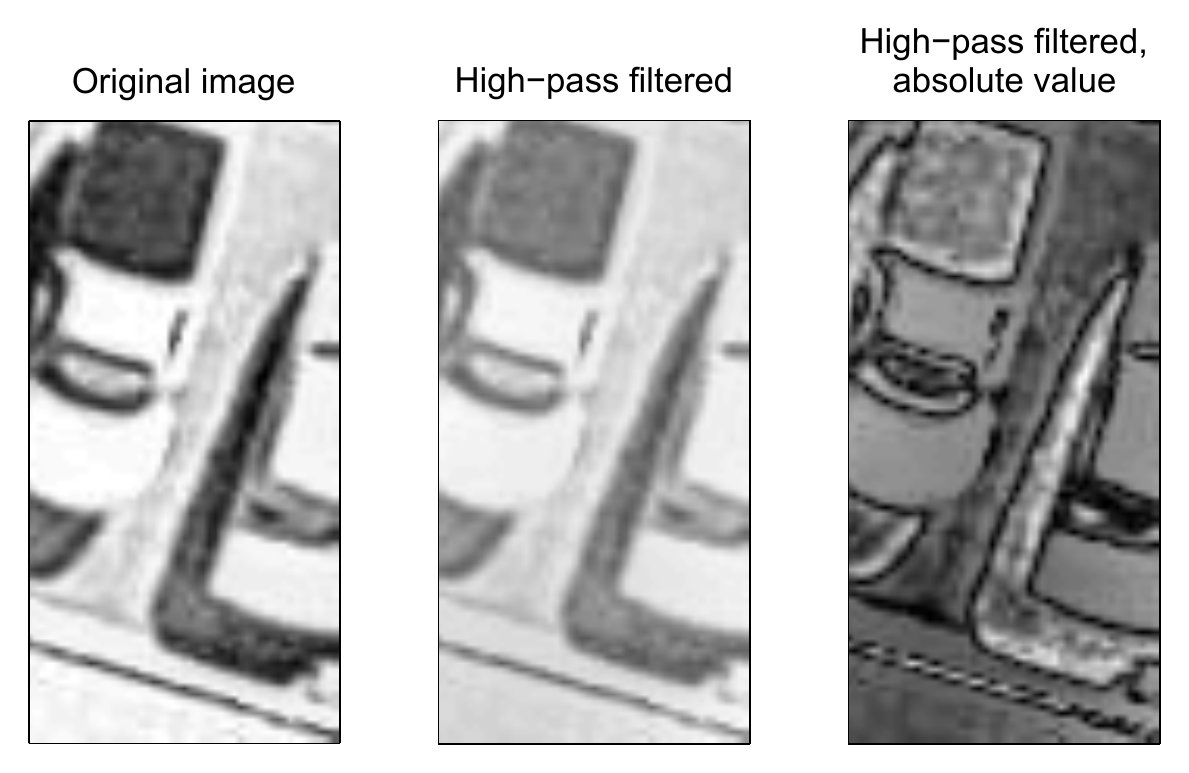}
  \caption{\small Image patch extracted from a raw input image (left), after high-pass filtering (centre), and after taking the absolute value of the high-pass filter output (right). }
  \label{f:representation}
\end{SCfigure}

Considering that illumination changes effect the most substantial appearance changes that we wish our representation to be invariant to, we focused our attention to various filters which preserve edge-like, high frequency information content in images. We experimented with high-pass filters \cite{AranCipo2006a,GangPradHolc2008}, quotient representations~\cite{Aran2009,Aran2013}, distance transformed edge maps \cite{LiuTuzeVeerChel2010,Aran2012,AranCipo2013}, and others \cite{Aran2012e}, with limited success. The representation that we found effective, and therefore which we adopt henceforth, is the absolute value of the high-pass filtered image:
{\small\begin{align}
  \zeta_i = \zeta(I_i)= \Big|~I_i - \big\{I_i \ast G(\sigma)\big\}~\Big|,
  \label{e:abshp}
\end{align}}
where $G(\sigma)$ is the isotropic 2D Gaussian kernel with the standard deviation `width' parameter $\sigma$, and $\ast$ denotes 2D convolution. The quality of registration agreement between two such quasi-illumination invariant images $\zeta_i$ and $\zeta_j$, geometrically transformed for the specified registration parameters, is then quantified using the normalized cross-correlation coefficient $\hat{\rho}(\Delta r_{i,j};\zeta_i,\zeta_j)$:
{\small\begin{align}
  \hat{\rho}(\Delta r_{i,j};\zeta_i,&\zeta_j) =
   \frac{ \sum_r \zeta_i(r)~\zeta_j(r+\Delta r_{i,j})} { \sqrt{\sum_{r} (\zeta_i(r))^2 \times \sum_r (\zeta_j(r))^2}}.
  \label{e:nxc}
\end{align}}
While the use of a high-pass filter ensures that the most significant responses occur around edge-like structures, taking its absolute value achieves invariance to the sign of the corresponding gradients, i.e.\ bright-dark vs.\ dark-bright interfaces. We will refer to this representation as ABS-HP. An example is shown in Fig~\ref{f:representation}.

\subsubsection{Computational considerations}\label{sss:computational}
The computation of the normalized cross-correlation coefficient $\hat{\rho}(\Delta r_{i,j};\zeta_i,\zeta_j)$ in \eqref{e:nxc} can be extremely slow -- if implemented `na\"{i}vely' in the image domain the number of computations is approximately $4 \times w \times h$, where $w$ and $h$ are the width and the height of an image in pixels. This is a potential bottleneck, as the value of the coefficient is needed for a different $\Delta r_{i,j}$ in each iteration of the maximization of the fitness function in \eqref{e:fitfn}. It is an attractive feature of the proposed framework that it lends itself to an efficient solution of this problem. Firstly, we employ the well-known fast Fourier transform-based pre-computation of the full cross-correlation matrix $\rho(\Delta r_{i,j}; \zeta_i,\zeta_j)$ \cite{ReddSrinChat1996}. Briefly, the cross-correlation between images $\zeta_i$ and $\zeta_j$, defined as:
{\small\begin{align}
\rho(\Delta r_{i,j}; \zeta_i,\zeta_j) = \sum_r \zeta_i(r)~\zeta_j(r+\Delta r_{i,j}) = \left( \zeta_i \ast \zeta_j \right) (\Delta r_{i,j}), \notag
\end{align}}
can be computed efficiently in the Fourier domain by exploiting the convolution theorem:
{\small\begin{align}
\rho(\Delta r_{i,j}; \zeta_i,\zeta_j) = \mathcal{F}^{-1}\left\{ \mathcal{F}(\zeta_i)^* \cdot \mathcal{F}(\zeta_j)  \right\}
\end{align}}
where $\mathcal{F}$ is the Fourier transform operator, $\cdot$ point-wise multiplication, and $(\ldots)^*$ complex conjugation. This solves the problem of computing the numerator in \eqref{e:nxc}. The computation is performed once and need not be repeated in each iteration, but rather the corresponding value looked-up. However, it is crucial that this value is  normalized by the denominator in \eqref{e:nxc}; the reason for this is that for different registration adjustments $\Delta r_{i,j}$, different parts of two images overlap. The omission of normalization could lead to an unfair bias towards large shifts because small overlapping image patches on average tend to look more alike. Thus, we pre-compute the normalization values using the integral image technique \cite{ViolJone2001a}. Specifically, we compute integral images for $\zeta_i \cdot \zeta_i$ and $\zeta_j \cdot \zeta_j$, which allows us quickly to obtain the values of all possible denominators in \eqref{e:nxc} in at most three elementary operations per denominator (since the overlapped area of an image always extends from one of its corners).

\subsection{Fitness function maximization}\label{ss:optimization}
Having pre-computed the full normalized cross-correlation matrix of the filtered images which are being registered, the fitness function introduced in \eqref{e:fitfn} can be readily maximized using the steepest ascent method; we initialize the process by setting $\forall i.~\Delta r_i=0$. The final issue we address here concerns the difficulty posed by what can be described as limited spatial influence of characteristic features extracted by our ABS-HP representation. Consider what happens when the initial misalignment between two images is large. Because our representation is based on a high-pass filter, the maximal filter responses are observed around edge-like structures. Since these structures are narrow, their responses do not have enough spatial reach to guide the optimization in the correct direction. While this problem does become much less noticeable when larger image sets are used, rather than the minimal set comprising two images, it is nonetheless beneficial for this potential pitfall to be avoided altogether. We achieve this by approaching the problem in a coarse-to-fine fashion. Specifically, observe that by varying the bandwidth of the high-pass filter used to extract our ABS-HP representation, it is possible to trade-off the breadth of spatial influence of a filter response, and its localization power. Thus, we start the registration process by using a wide-band high-pass filter, and follow that by progressively narrower band filters as convergence at each level is detected. The particular filters we used in our experiments have the values of 40, 20, 8, and 3 pixels for the parameter $\sigma$ in \eqref{e:abshp}.

\section{Evaluation}
In this section we describe our evaluation of the proposed method, and report and discuss its performance in the context of the current state-of-the-art. We begin by describing our data set and evaluation protocol, follow with a presentation of a comprehensive set of performance statistics, and finish off with an analysis of our results and their significance.

\subsection{Data}\label{ss:data}
As reviewed in detail in Sec~\ref{s:intro}, the existing literature on the registration of aerial imagery is void of any set-based approaches, the present paper being the pioneering work in the area. It is an unsurprising consequence of this that there are no public data sets suitable for the evaluation of set-based algorithms so we collected a novel data set ourselves.

\begin{figure}[t]
  \centering
  \includegraphics[width=0.45\textwidth]{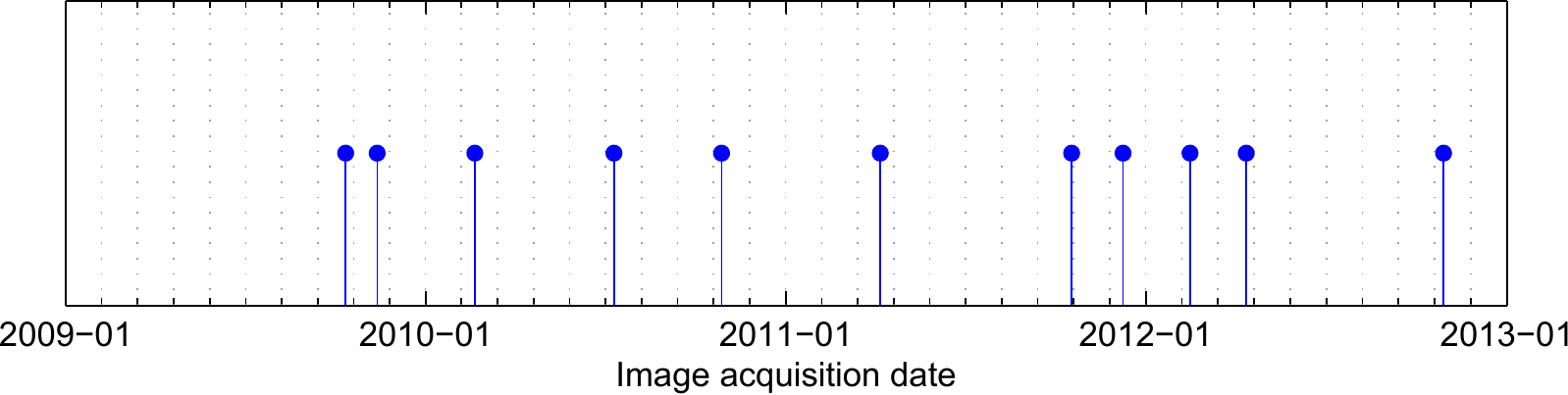}
  \caption{\small The distribution of acquisition dates for a typical image set in our data. It can be readily seen that the acquisition was not performed at regular intervals: in some instances images are re-acquired after only a month, while in others several months pass. }
  \label{f:dates}
\end{figure}

Our data set comprises 10 image sets, each set containing 10 images acquired at different, non-uniformly distributed dates, as illustrated in Fig~\ref{f:dates}. This data was manually downloaded using the freely accessible web portal provided by Nearmap Ltd. Nearmap has developed technology for rapid acquisition of high resolution aerial imagery, which allows frequent re-imaging of large areas. The most time demanding task in their pipeline concerns the registration and stitching of aerial images (tiles) to form a continuous representation. Registration is performed using a combination of manual input and state-of-the-art commercial software; thus, different images within the same image set in our data correspond to approximately the same land area. Both GPS and image data are used in the registration process, the latter being based on a robust alignment of local features (the exact details of the registration algorithm are proprietary and as such were not disclosed to us in entirety).

\subsection{Protocol}\label{ss:protocol}
We first obtained an estimate of the ground truth by manually labelling images. Specifically, the optimal registration parameters were estimated from correspondences between selected characteristic image loci. Loci physically lying on the ground plane were consistently chosen to avoid the problem of varying viewpoint from which images are obtained, and which can significantly change the perspective from which different surfaces are seen (e.g.\ house walls or roofs).

After obtaining an estimate of the ground truth, we conducted three baseline experiments. Firstly, we assessed the quality of registration performed by Nearmap. In addition, we evaluated two popular registration approaches from the literature: (i) using SURF \cite{BayEssTuytGool2008} feature correspondences (similar to \cite{Aran2012d} using SIFT), and (ii) the state-of-the-art ARRSI algorithm \cite{WongClau2007}, specifically tailored to aerial images, which is also sparse in nature and uses phase congruency-based control points. Feature matching in both cases is performed robustly using RANSAC.

\subsection{Results and discussion}\label{ss:results}
Using our ground truth labelling, we were able to estimate that the average registration error of the proprietary method employed by Nearmap is approximately 18~pixels. Much like most of Nearmap's imagery, our data was acquired in the resolution for which one pixel width corresponds to 7.5~cm on the ground. Thus, the average misalignment of two images considered to show the same patch of land by Nearmap is about 1.35~m. This error is more than sufficiently large to limit the powers of subsequent processing for image understanding; to give an example, this may be the detection of permanent structural change (e.g.\ solar panel installation, well drilling etc), which is of major interest to local councils and governments. It is insightful to notice that while the standard deviation of the mean misalignment error between sets was found to be very small indeed (0.75~pixels, or 5.6~cm on the ground), the mean deviation within a set was far larger (24.6 pixels, or 1.85~m); please see Table~\ref{t:res}. This strongly supports one of the premises of our work: that the primary challenge is not posed by the content of the aerial scene (i.e.\ the structures in it), but rather the changes that the scene exhibits over time (illumination being the most substantial one).

\begin{table}
  \centering
  \renewcommand{\arraystretch}{1.2}
  \small
  \caption{\small  Baseline performance (all statistics are in pixels).  }
  \vspace{5pt}
  \begin{tabular}{l|ccc}
    \Hline
    Method             & Nearmap               & ARRSI                  & SURF\\
    \hline
    Mean error         & 18.0                  & 248.5                  & 140.9\\
    Deviation          & \multirow{2}{*}{0.75} & \multirow{2}{*}{77.2}  & \multirow{2}{*}{110.9}\\[-2pt]
    (between sets)     &    &  &\\
    Deviation          & \multirow{2}{*}{24.6} & \multirow{2}{*}{139.8} & \multirow{2}{*}{266.9}\\[-2pt]
    (within a set, mean) &    &  &\\
    \Hline
  \end{tabular}
  \label{t:res}
\end{table}

We next turn our attention to the two baseline methods from the literature. As the summary in Table~\ref{t:res} clearly shows, both of these performed very poorly on our data. Not only did neither of the methods manage to improve on the original registration by Nearmap, both of them increased the average registration error by approximately an order of magnitude. While perhaps surprising at first, this finding is readily explained following a more in-depth examination of the results. Specifically, in both cases observe the extremely large deviation of the misalignment error within an image set -- while in the case of some image pairs registration was highly successful (error of $\approx 4$ pixels), in other cases unreliable interest point correspondences resulted in grossly inaccurate results (errors in excess of 400~pixels). Indeed, as stated in Sec~\ref{ss:local}, this is consistent with our experiments using a variety of interest point descriptors -- while highly successful in the registration of images acquired in similar illumination conditions, even in the presence of different small transient objects, sparse feature-based methods exhibit a dramatic drop in performance as illumination conditions change. We found them to lack sufficient robustness to deal with the challenges in real-world images such as those in our data set.

\begin{figure}[thp]
  \centering
  \subfigure[]{\includegraphics[height=0.16\textwidth]{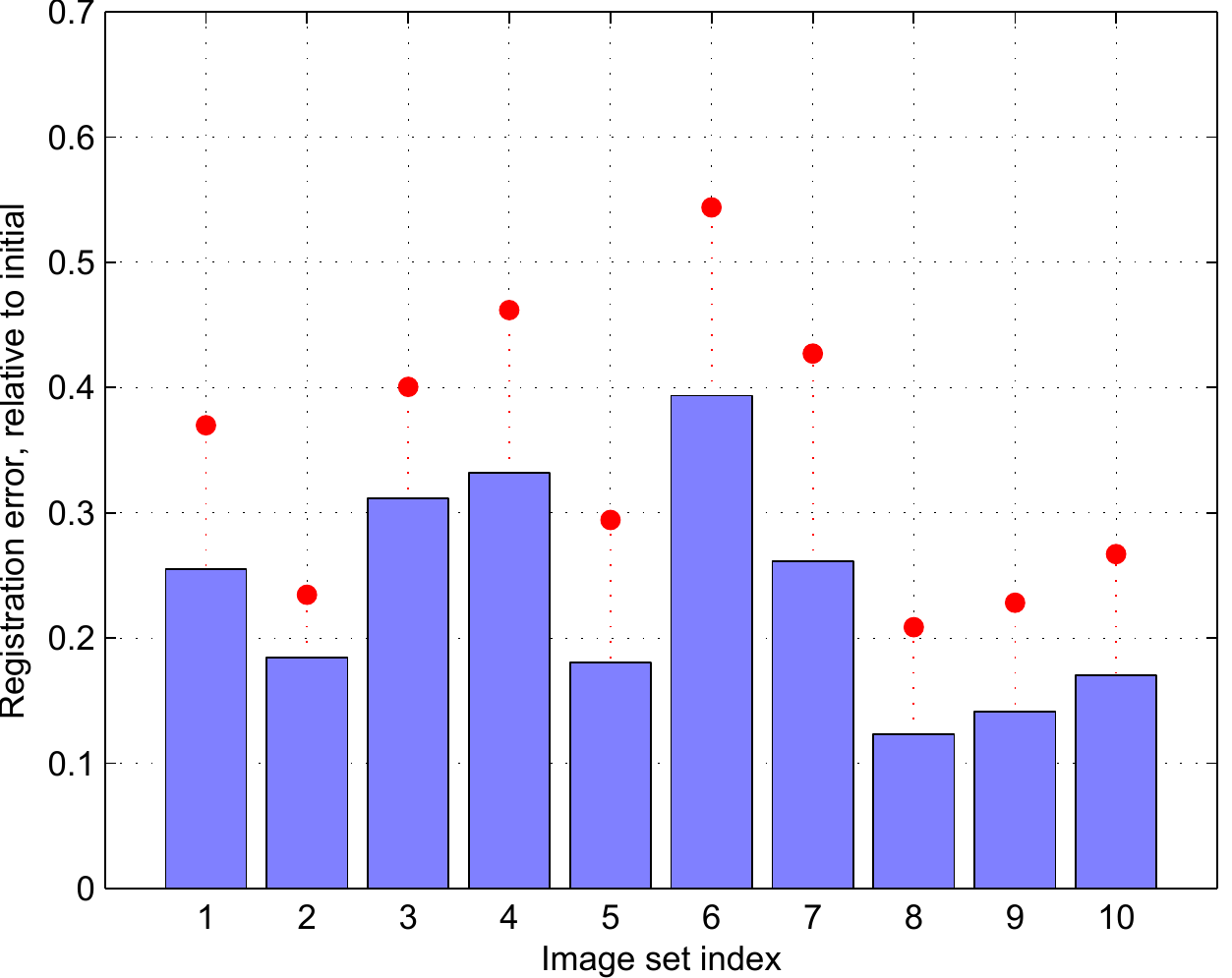}\label{f:error}}~~~~~~
  \subfigure[]{\includegraphics[height=0.16\textwidth]{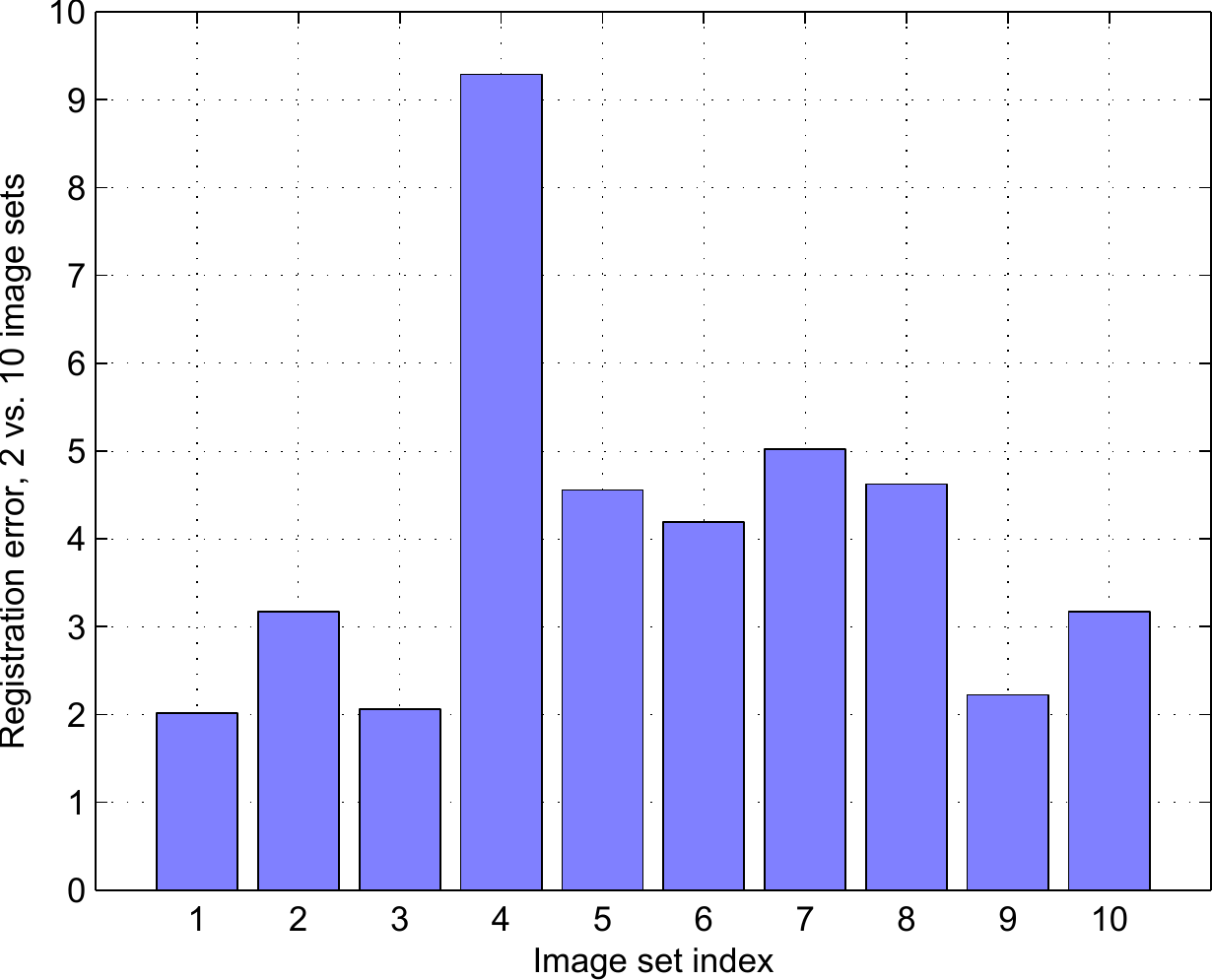}\label{f:error2vs10}}
  \vspace{-8pt}
  \caption{\small (a) Mean pair-wise registration error (blue bars) per image set obtained with the proposed method, relative to the initial error. In all cases the error reduction is dramatic, averaging 75.8\%. The red stems extending from the blue bars show the standard deviation of the relative registration error within each of the 10 image sets. (b) The reduction in the mean registration error per image set using 10-image sets relative to 2-image sets. In all cases the error is reduced at least to half, and 3.2 times on average. }
\end{figure}

Lastly, we present the results obtained using the proposed method. For all sets we found that our method substantially decreased Nearmap's registration error. On average, the reduction was 75.8\%, resulting in the average absolute image error of only 4.4~pixels, or 32.7~cm on the ground. A plot detailing the performance of the method is shown in Fig~\ref{f:error}. While our method too exhibited some variation across different sets, even in the worst case (number 6) the error was reduced by over 60\%. This demonstrates the achievement of our first goal of developing both a more accurate, and a more robust registration method, than the state-of-the-art used commercially, or indeed described in the academic literature.

A major premise of our work, and a methodological novelty, pertaining to the joint co-registration of aerial image sets rather than image pairs is substantiated by the data shown in Fig~\ref{f:error2vs10}. This plot compares the reduction in the mean registration error (relative to Nearmap's baseline error) when our method is applied to minimal 2-image sets (i.e.\ subsets of the original sets), and when all the available data (10 images) is used instead. Indeed, it can be readily seen that the error is consistently reduced in the case of all sets. Even in worst case (set number~1) the reduction is over two-fold, while in the best case it is more than nine-fold (set number~4).

The variation in the benefit -- that is, the reduction in the registration error -- across sets corresponding to different land areas obtained by using 10 as opposed to 2 images led us to investigate this specific aspect of our results in greater detail. By examining the variation in the average registration error as the size of an image set is gradually increased we noticed that the variation observed in Fig~\ref{f:error2vs10} primarily emerges as a consequence of the variation in the registration error obtained for the minimal set size (i.e.\ pair-wise registration)~\cite{AranPhamVenk2015c}. Already for sets of size 3 the variation (absolute, as well as relative) across different sets is much reduced. This finding too strongly supports the premise that the constraints which can be extracted from the use of more than two images are a powerful source of information which can be harnessed to increase the accuracy and reliability of registration in the presence of large appearance changes.

Lastly, a summary of the computational cost statistics is given in Table~\ref{t:time}. As expected, the average time for registration per image achieved using the proposed method increases somewhat with the increase in the set size due to the greater complexity of the constraints graph. Nonetheless, in all cases our method's computational cost was significantly lower than that of either of the state-of-the-art methods.

\begin{table}[bt]
  \centering
  \renewcommand{\arraystretch}{1.2}
  \small
  \caption{\small Computational cost comparison. ARRSI and the SURF-based methods were implemented primarily in C, with a Matlab `wrapper', while the proposed method was implemented fully in Matlab. The estimates are averages of 100 executions ran in Matlab 7 on an AMD Phenom II X4 965 processor with 8GB RAM. }
  \vspace{5pt}
  \begin{tabular}{l|ccccc}
    \Hline
    Method                        & \multicolumn{3}{c}{Proposed} & ARRSI & SURF\\
    \hline
    Set size                      & 10  & 5   & 3                &  n/a  & n/a\\
    \hline
    Registration time  & \multirow{2}{*}{5.6} & \multirow{2}{*}{3.8} & \multirow{2}{*}{3.9}              & \multirow{2}{*}{17.5}                  & \multirow{2}{*}{22.5}\\[-2pt]
    (s per image)        & & &               &                   &\\
    \Hline
  \end{tabular}
  \label{t:time}
\end{table}

\section{Summary and conclusions}\label{s:summary}
In this paper we introduced a novel method for the registration of aerial images. Unlike previous work which considered either pair-wise image-to-image or image-to-map registration, our approach jointly registers an entire set of images of approximately the same area but acquired at different times.  We formulated the joint registration problem as an optimization scheme. Set-based registration was built upon simple pair-wise registrations which are mutually constrained by means of a connectivity graph. We showed how this graph can be constructed automatically, using a combination of two rules, one proximal, the other distal in the Euclidean image space. Using a novel data set suitable for the evaluation of set-based aerial image registration algorithms, we demonstrated that the proposed approach significantly outperforms the current state-of-the-art both in terms of accuracy and reliability, as well as speed. Amongst other possible directions for improvement, our future work will investigate the use of colour invariants~\cite{Aran2012b}.

\small
\bibliographystyle{named}
\bibliography{../../../my_bibliography}

\end{document}